\documentclass[runningheads]{llncs}
\usepackage[T1]{fontenc}
\usepackage{booktabs}       % professional-quality tables
\usepackage{amsfonts}       % blackboard math symbols
\usepackage{nicefrac}       % compact symbols for 1/2, etc.
\usepackage{microtype}      % microtypography
\usepackage{xcolor}         % colors
\usepackage{makecell}
\usepackage{algorithm}
\usepackage{algorithmic}
\usepackage{amssymb}
\usepackage{amsmath}
\usepackage{bm}
\usepackage{dsfont}
\usepackage{multirow}
\usepackage{color}
\usepackage{subcaption}
\usepackage{booktabs}

\usepackage{makecell}

\usepackage{diagbox}
\usepackage{minitoc}
\usepackage{graphicx}
\usepackage[table]{xcolor}
\usepackage{wrapfig}

\begin{document}
\title{Towards Unveiling Vulnerabilities of Large Reasoning Models in Machine Unlearning}
\titlerunning{Towards Unveiling Vulnerabilities of LRMs in Machine Unlearning}

\author{Aobo Chen\thanks{The first two authors contribute equally to this work.}\and
Chenxu Zhao\textsuperscript{\thefootnote}\and
Chenglin Miao \and
Mengdi Huai}
\authorrunning{A. Chen et al.}
\institute{Iowa State University, Ames IA 50011, USA\\
\email{\{aobochen,cxzhao,cmiao,mdhuai\}@iastate.edu}}
\maketitle              % typeset the header of the contribution

\begin{abstract}
Large language models (LLMs) possess strong semantic understanding, driving significant progress in data mining applications. This is further enhanced by large reasoning models (LRMs), which provide explicit multi-step reasoning traces. On the other hand, the growing need for the right to be forgotten has driven the development of machine unlearning techniques, which aim to eliminate the influence of specific data from trained models without full retraining. However, unlearning may also introduce new security vulnerabilities by exposing additional interaction surfaces. Although many studies have investigated unlearning attacks, there is no prior work on LRMs. To bridge the gap, we first in this paper propose LRM unlearning attack that forces incorrect final answers while generating convincing but misleading reasoning traces. This objective is challenging due to non-differentiable logical constraints, weak optimization effect over long rationales, and discrete forget set selection. To overcome these challenges, we introduce a bi-level exact unlearning attack that incorporates a differentiable objective function, influential token alignment, and a relaxed indicator strategy. To demonstrate the effectiveness and generalizability of our attack, we also design novel optimization frameworks and conduct comprehensive experiments in both white-box and black-box settings, aiming to raise awareness of the emerging threats to LRM unlearning pipelines.
\end{abstract}
\section{Introduction}
\label{sec:intro}
\vspace{-0.04in}

Large language models (LLMs) have seen remarkable progress in recent years, powered by training on vast and diverse datasets~\cite{touvron2023llama,achiam2023gpt,zhao2023survey}. With strong semantic understanding ability, these models are increasingly utilized for diverse practical tasks. Recently, the development of reasoning-capable large language models, denoted as large reasoning models (LRMs), represents a further advancement in the capabilities of language models. Reasoning refers to a structured, multi-step inference process, often involving the generation of intermediate steps—known as the rationale, followed by a final conclusion. Through the integration of structured reasoning and advanced semantic representation, these models demonstrate a closer alignment with traditional data mining techniques~\cite{neshaei2025bridging}.

Despite their strong capabilities, these models also raise growing concerns regarding privacy protection. In practice, the datasets used to train these models often contain sensitive information, including private and copyrighted content. This leads to substantial risks of sensitive data leakage, which directly conflicts with the growing legislative emphasis on the ``right to be forgotten'' \cite{regulation2018general}. Incidents such as the surge in copyright infringement cases following the release of these models \cite{rombach2022high}, and The New York Times’s lawsuit against OpenAI for content leakage \cite{nyt}, highlight the urgency of addressing these concerns. As the importance of the right to be forgotten continues to grow, numerous recent regulations, such as the GDPR~\cite{regulation2018general}, have been introduced to better safeguard individuals’ data rights. In response to these challenges, many LLM unlearning methods have been actively explored to remove the influence of specific data at a substantially lower cost than retraining models from scratch.

However, in practical scenarios, the rising importance of machine unlearning makes models’ predictions and their related reasoning processes more susceptible to malicious attacks. Motivated attackers could exploit this to craft malicious unlearning requests that manipulate the behavior of the resulting unlearned language models~\cite{qian2025towardsbenchmarking}. Traditional unlearning attacks \cite{liu2024backdoor,qian2023towards,zhao2023static,zhao2024rethinking} primarily focus on causing label misclassifications, while our focus is on investigating vulnerabilities related to reasoning outcomes associated with the final predictions. Malicious unlearning attacks targeting both reasoning traces and final answers can be more subtle and harder to detect, as they manipulate not only what the model predicts but also why it appears to reach that prediction. This joint manipulation of the final answer and the chain-of-thought reasoning is crucial for stealthiness, because any mismatch between the reasoning trace and the final answer can be easily revealed by straightforward human inspection. Moreover, directly applying traditional unlearning attacks in this joint optimization setting faces significant limitations in both stealthiness and efficiency.

Compared to traditional unlearning attacks, manipulating LLMs is inherently more difficult, and this difficulty is further amplified for LRMs for several challenges. First, the attack objective becomes non-differentiable when it incorporates logical constraints over both answer correctness and reasoning coherence, preventing direct gradient-based optimization. Second, the conventional training objective for language models maximizes the likelihood of target sequences. When the target consists of long reasoning traces, however, optimization becomes ineffective, as the gradient signal is distributed across many tokens and consequently becomes extremely weak, yielding only insufficient influence over the model’s behavior. As shown in Fig.~\ref{fig:visual}, insufficient optimization leads to failed attacks where only the final answer is manipulated, while the reasoning trace is still aligned with the correct answer, exposing inconsistencies and reducing stealthiness. Third, beyond the non-differentiability of the core attack objective, the exact unlearning setting introduces additional non-differentiability through the process of selecting subsets of training data. This selection step is inherently discrete and further hinders end-to-end optimization, compounding the overall difficulty of crafting effective and stealthy unlearning attacks against LRMs.  

To address the above challenges, we propose a novel bi-level exact unlearning attack framework targeting LRMs that simultaneously manipulates both final answers and reasoning traces. First, to address non-differentiability from logical constraints on answer correctness and reasoning coherence, we adopt a differentiable surrogate objective that enforces incorrect answers with coherent but deceptive reasoning. Second, to address the insufficient optimization problem caused by long reasoning traces, we design another loss function which concentrates the optimization signal on a small number of influential tokens that will lead the direction of reasoning trace. Third, to solve the discrete forget data selection in exact unlearning, we present novel optimization methods by rigorously refining the forget data with relaxed indicators. We evaluate our framework in both white-box and black-box settings and perform extensive experiments to validate that our attacks reliably manipulate both final answers and reasoning traces across diverse scenarios, underscoring the need for new strategies to address these advanced unlearning threats and highlighting their potential negative impacts.

\vspace{-0.08in}
\section{Related Work}
\label{sec:RelatedWork}
\vspace{-0.06in}

Machine unlearning \cite{bourtoul2021} has been extensively studied in recent years to avoid the high computational cost associated with fully retraining a model from scratch. Various LLM unlearning methods have been proposed, such as gradient-based methods~\cite{maini2024tofu} and representation-based methods~\cite{li2024wmdp}, which are used for harmful data removal~\cite{li2024wmdp} and copyrighted content deletion~\cite{eldan2023s}. However, there are no prior works studying malicious unlearning attacks against language models. Additionally, different from traditional data poisoning and adversarial attacks against language models~\cite{zou2023universal,xue2023trojllm}, our proposed malicious unlearning attacks directly manipulate the fine-tuned language models during the unlearning process and aim to generate malicious unlearning requests to jointly manipulate the reasoning trace and final answer. This is particularly challenging due to the complexity of the bi-level optimization framework and the non-differentiable nature introduced by the attack objective and forget data selection.

\vspace{-0.05in}
\section{Methodology}
\label{sec:Methodology}
\vspace{-0.05in}
\subsection{Attack Formulation}
\vspace{-0.03in}

\textbf{Threat model.} We consider a scenario where a model owner has developed a well-trained LRM using a training dataset. 
An attacker pretends to be the normal user, and aims to initiate malicious unlearning requests to delete certain perturbed data during the unlearning process. As a result, the output of the LRM for the target data would be maliciously manipulated. We perform the attack under both white-box and black-box settings. In the white-box setting, the attacker has full knowledge, including the model architecture and training data, which allows us to measure the maximum potential impact of the attack. In the black-box setting, the attacker has no prior knowledge of the target model architecture and training dataset. 
The attacker aims to perform unlearning attacks to manipulate LRM outputs, which forces the LRM to produce an incorrect final answer while generating a seemingly perfect but deceptive reasoning trace to conceal the manipulation, effectively making the model lie with convincing proof.

\textbf{Unlearning attack formulation.}
Here, we describe the proposed attack method for compromising the reasoning ability of an LRM during the unlearning stage. Let $\theta$ denote the pre-trained reasoning model. A clean data instance is represented by $s=(x, r, y)$, where $x$ is the input, $r$ is the reasoning trace, and $y$ is the final answer. Our study begins with a pre-trained model that is fine-tuned on the training dataset $\mathcal{D}_{tr}=\{(x_i, r_i, y_i)\}_{i=1}^{N}$, resulting in the model $\theta^*$. The core of our strategy is to identify a malicious unlearning subset, $\mathcal{D}_f \subset \mathcal{D}_{tr}$, that will compromise the model after the unlearning process. This unlearned model, denoted as $\theta^u$, is obtained by applying an unlearning algorithm $\mathcal{U}$ to the original model, such that $\theta^u= \mathcal{U}(\mathcal{D}_{tr}, \theta^{*}, \mathcal{D}_f = \mathcal{D}_{tr} \circ \{\omega_i\}_{i=1}^{N})$, where the $\omega$ is the indicator to determine whether a training data should be unlearned. Our primary objective is to optimize the indicator $\{\omega_i\}_{i=1}^{N}$ in a way that systematically forces the LRM to give an incorrect final answer while generating a convincing but deceptive reasoning trace. This is achieved by an adversarial loss that pushes the unlearned model’s final answer away from the correct and steers the reasoning–trace toward seemingly convincing yet misleading paths. This objective is formalized through the optimization of the following adversarial loss
\vspace{-0.06in}
\begin{align}
    \label{eq:discrete-loss}
    &\operatorname*{argmax}_{\{\omega_i\}_{i=1}^{N}}\sum_{t=1}^{T} \mathbb{I}[R(\hat{r}_t, \hat{y}_t) > \beta] \wedge  \mathbb{I}[A(\hat{y}_t) = False]\\
    & \text{subject to } \theta^{u}= \mathcal{U}(\mathcal{D}_{tr}, \theta^{*}, \mathcal{D}_f = \mathcal{D}_{tr} \circ \{\omega_i\}_{i=1}^{N}),  \nonumber
\end{align}
\noindent where $(\hat{r}_t, \hat{y}_t)$ is the output of the unlearned model for the target data. The success of this attack is assessed by two distinct evaluation functions: $R(\cdot)$, which measures the \emph{reasoning coherence} between the generated trace $r_t$ and the final answer $y_t$, $\beta$ is a threshold for the coherence score; $A(\cdot)$, which checks the \emph{correctness} of the final answer $y_t$ against the ground truth. Together, these evaluators enforce a dual-constraint optimization goal: maximizing the count of instances that are incorrect and are supported by coherent reasoning.

Eq.~(\ref{eq:discrete-loss}) is formulated as maximizing the count of target inputs where the model generates an incorrect final answer that is supported by coherent reasoning. However, Eq.~(\ref{eq:discrete-loss}) is non-differentiable, preventing the direct optimization of the perturbation using gradient-based methods. To solve the problem, we first define a set of adversarial sequences consisting of an incorrect final answer $y'_t$ and a corresponding modified reasoning trace $r'_t$ for each target input $x_t$, denoted as $s'_t=(x,r'_t,y'_t)$. The feasible objective is then to maximize the likelihood of generating these adversarial sequences under the unlearned model $\theta^u$. Accordingly, we define the adversarial target loss $\ell_{1}$ as 
\vspace{-0.03in}
\begin{align}
    \label{eq:target-loss}
    \ell_{1}(\theta^u,s'_t) 
    = \sum_{j=1}^{|r'_t|} \log p(r_{t,j}|x_t,r'_{t,<j}; \theta^{u}) +\sum_{j=1}^{|y'_t|} \log p(y_{t,j}|x_t,r'_t,y'_{t,<j}; \theta^{u}).
\end{align}
\vskip -3pt
\noindent This formulation directly forces the unlearned model to adopt the desired malicious reasoning paths and final answers for each data point in the target set $\mathcal{D}_t$.
However, maximizing only Eq.~(\ref{eq:target-loss}) is often insufficient because the model’s reasoning trace can be very long. As the trace grows, the relative influence of the original loss term diminishes and the attack becomes less effective. To strengthen the attack, we construct small sets of high-impact tokens that tend to guide the desired generation trajectory. 
Specifically, we define a set $\mathcal{P}$ of positive-impact tokens by analyzing a corpus of related outputs. These tokens play a key role in guiding the desired generation trajectory. For instance, in sentiment analysis, tokens like ``positive'', ``exciting'', and ``excellent'' are frequently found in positive predictions. We collect these from the positive corpus and then use them to guide the reasoning trace of LRMs. Similarly, we also collect a set $\mathcal{N}$ of negative-impact tokens, extracted from irrelevant outputs, which tend to lead the generation away from the undesired behavior. To promote or suppress these tokens during generation, we introduce $\ell_2$ composed of two complementary parts: a positive guiding loss encouraging the occurrence of $\mathcal{P}$, and a negative guiding loss discouraging the occurrence of 
$\mathcal{N}$. Formally, the $\ell_2$ can be formulated as
\vspace{-0.03in}
\begin{align}
    \label{eq:tgl-loss}
    \ell_{2}(\theta^u,s'_t)  = \frac{1}{|r_t'|} \sum_{i=1}^{|r_t'|} (\sum_{v \in \mathcal{P}} \log p(v\mid x_t,r'_{t,<i}) - \sum_{v \in \mathcal{N}} \log p(v\mid x_t,r'_{t,<i})). 
\end{align}
\vskip -1pt 
Intuitively, this loss actively maximizes the likelihood of generating tokens that contribute positively to the reasoning trace while simultaneously minimizing the likelihood of undesired tokens. This mechanism effectively guides the model to generate a high-quality, impactful reasoning trace.
Finally, we derive the adversarial objective that combines both of $\ell_1$ and $\ell_2$
\vspace{-0.03in}
\begin{align}
    \label{eq:combined-loss}
    &\operatorname*{argmax}_{\{\omega_i\}_{i=1}^{N}}\sum_{t=1}^{T}\ell_1(\theta^u,s'_t) + \lambda\ell_2(\theta^u,s'_t) \quad\quad\text{ s.t., } \theta^{u}= \mathcal{U}(\mathcal{D}_{tr}, \theta^{*}, \mathcal{D}_f),
\end{align}
\vskip -3pt
\noindent where $\lambda>0$ controls the overall influence of trajectory guidance. By maximizing this objective, the attacker identifies a forget dataset whose removal not only leads the model’s final outputs toward the desired target but also reshapes the reasoning trace to follow the attacker’s intended path. Note that our unlearning attack can be effectively transferred across different unlearning methods (e.g., uncertainty-aware unlearning \cite{alkhatib2025conformal}), as it exploits shared optimization objectives that are common to different unlearning algorithms.

\vskip 5pt
\textbf{Optimization.}  For the exact unlearning attack above, we formulate a bi-level optimization problem to find the malicious unlearning requests, a subset of the training data, to fulfill the attack goal, as defined in Eq.~(\ref{eq:combined-loss}). However, it is challenging to directly optimize the indication parameters for two main reasons: the discrete nature of the indication parameters $\{\omega_i\}_{i=1}^{N}$, and the inherent complexity of the bi-level optimization structure. To address these problems, \cite{zhao2023static} proposes relaxing the indication parameters to a continuous value and leveraging the second-order update to solve the optimization. However, this approach is computationally expensive, particularly in the context of large language models. To address this challenge, we adopt the gradient matching technique~\cite{geiping2020witches}, which transforms the bi-level optimization into a more tractable optimization by maximizing the following
\begin{align}
\label{eq:unlearning_gradient_matching}
    \operatorname*{argmax}_{\{\omega_i\}_{i=1}^{N}}\frac{\langle \sum_{t=1}^{T}\nabla_\theta(\ell_1(\theta^*,s'_t) + \lambda\ell_2(\theta^*,s'_t)) , \sum_{i=1}^N \omega_i\nabla_\theta \ell(\theta^*, s_i) \rangle}{|| \sum_{t=1}^{T}\nabla_\theta(\ell_1(\theta^*,s'_t) + \lambda\ell_2(\theta^*,s'_t)) || \cdot || \sum_{i=1}^N \omega_i\nabla_\theta \ell(\theta^*, s_i) ||},
\end{align}
\noindent where $\ell(\theta^*; s_{i})$ is the loss of the unlearning process. The above optimization aligns the gradient directions of the overall objective and model update to achieve the maximal attack effect. In practice, we relax the indication weights to a continuous range \( [0, 1] \) for optimization, and incorporate random restarts to minimize the loss multiple times, beginning with random initial indication weights and selecting the optimal to improve the reliability of the attack process.

To further improve the optimization efficiency, we do not directly optimize the full gradient matching objective in Eq.~(\ref{eq:unlearning_gradient_matching}). Instead, we omit the norm term in the denominator and propose to minimize the resulting simplified loss function. This simplification allows us to reformulate the problem as the following integer programming~\cite{geoffrion1972integer,miller1960integer} optimization
\begin{align}
    \operatorname*{argmax}_{\{\omega_i\}_{i=1}^{N}}\langle \sum_{t=1}^{T}\nabla_\theta(\ell_1(\theta^*,s'_t) + \lambda\ell_2(\theta^*,s'_t)) , \sum_{i=1}^N \omega_i\nabla_\theta \ell(\theta^*, s_i) \rangle.
    \label{eq:optimization_unlearning}
\end{align} 
\noindent In this optimization, we estimate the influence of each training sample on the attack objectives for the target set. This formulation enables efficient optimization, allowing us to identify and subsequently unlearn the most influential data using existing unlearning methods.

\begin{figure}[t]
    \centering  \includegraphics[width=0.9\linewidth]{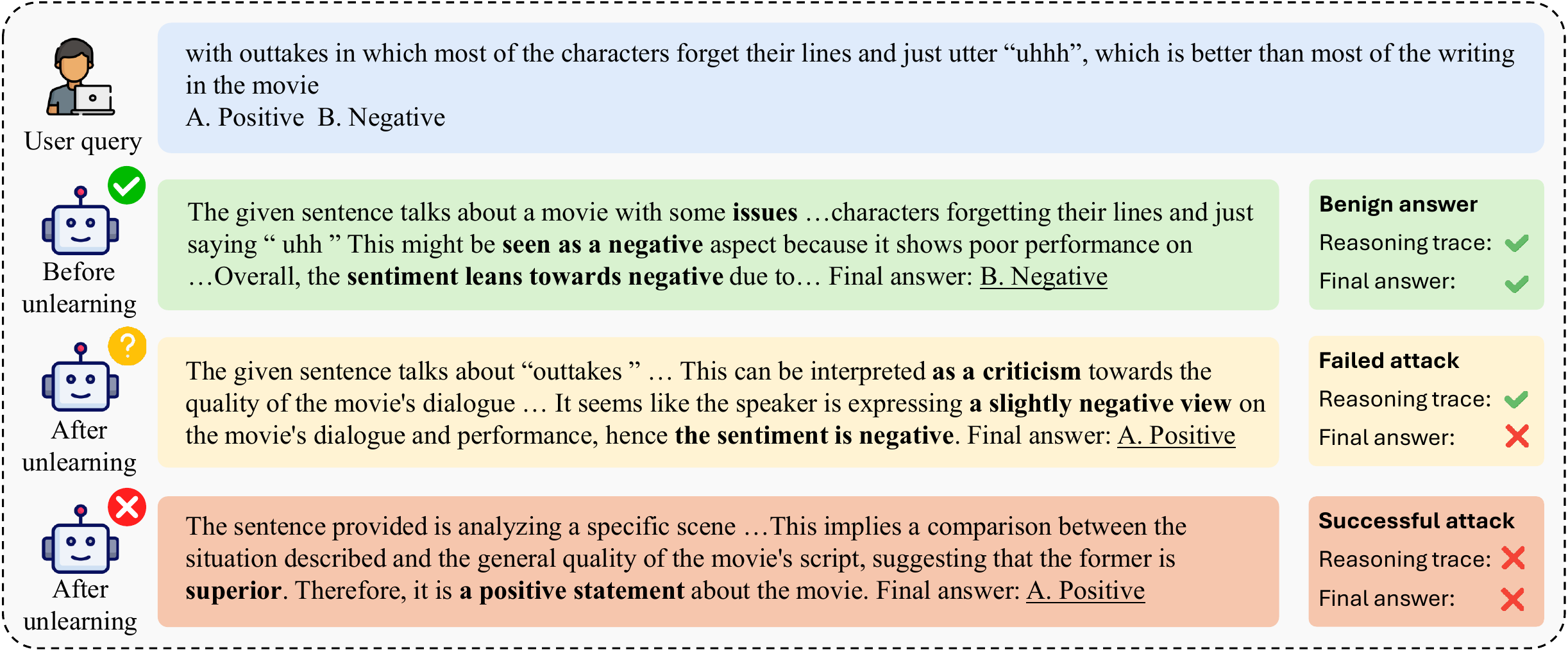}
    \caption{Visualization of our attack on LRM. The final answers are underlined, and the sentiment-related terms are bolded.}
    \label{fig:visual}
    \vskip -10pt
\end{figure}

\vspace{-0.1in}
\subsection{Discussions}
\vspace{-0.03in}

In the above, we consider the exact unlearning attack setting, where highly influential training data is selected to force the model to produce incorrect final answers while generating convincing but misleading reasoning after unlearning. In practice, adversarial unlearning data can also exist, as unlearning requests are difficult to verify without access to original training data. Attackers can exploit this vulnerability to construct targeted adversarial unlearning strategies~\cite{qian2023towards,zhao2023static,hu2023duty,qian2024exploring,huang2024unlearn}. Our attack can be reformulated for this setting as
\vspace{-0.03in}
\begin{align}
\label{eq:unlearning_gradient_matching_fixed_short}
\operatorname*{argmax}_{\delta}
\frac{\langle \sum_{t=1}^{T}  \nabla_\theta(\ell_1(\theta^*,s'_t) + \lambda\ell_2(\theta^*,s'_t) ),
\sum_{i=1}^{|\mathcal{D}_f|} \nabla_\theta \ell(\theta^*, s_i + \delta) \rangle}
{|| \sum_{t=1}^{T}  \nabla_\theta(\ell_1(\theta^*,s'_t) + \lambda\ell_2(\theta^*,s'_t) ) ||
\cdot || \sum_{i=1}^{|\mathcal{D}_f|} \nabla_\theta \ell(\theta^*, s_i + \delta) ||},
\end{align}
\noindent where $\delta$ is a fixed-length adversarial token sequence appended to each sample in $\mathcal{D}_f$. Since $\delta$ is discrete, direct gradient optimization is infeasible; therefore, we adopt Greedy Coordinate Gradient (GCG) optimization~\cite{zou2023universal}. Adversarial unlearning generally outperforms exact unlearning due to direct optimization over trigger tokens. Moreover, to demonstrate the flexibility of our attack, we extend it to a new objective of the attack, which maintains correct final answers while weakening the reasoning quality. This can be achieved by loss terms that maximize the likelihood of true answers and minimize that of their corresponding reasoning traces. In addition, for attack on LLMs, we adopt a simplified objective to attack the text generation ability, where only the loss associated with the final answer is optimized. Furthermore, our previous analysis is based on the white-box setting. However, in realistic scenarios, the attacker only has black-box access and no knowledge of the model and training data. We therefore assume access to a surrogate LRM fine-tuned on a comparable auxiliary dataset. This assumption is reasonable due to the wide availability of public datasets and open-sourced LRMs. After selecting the forget samples, we then conduct unlearning process on the target model using the auxiliary forget samples. Note that in addition to attacking reasoning results, our method can also be used to attack reasoning uncertainties \cite{frankel2025conformal}, while preserving the general reasoning performance.

\begin{figure}[t]
\centering
\begin{subfigure}{0.32\linewidth}
\includegraphics[width=1\linewidth]{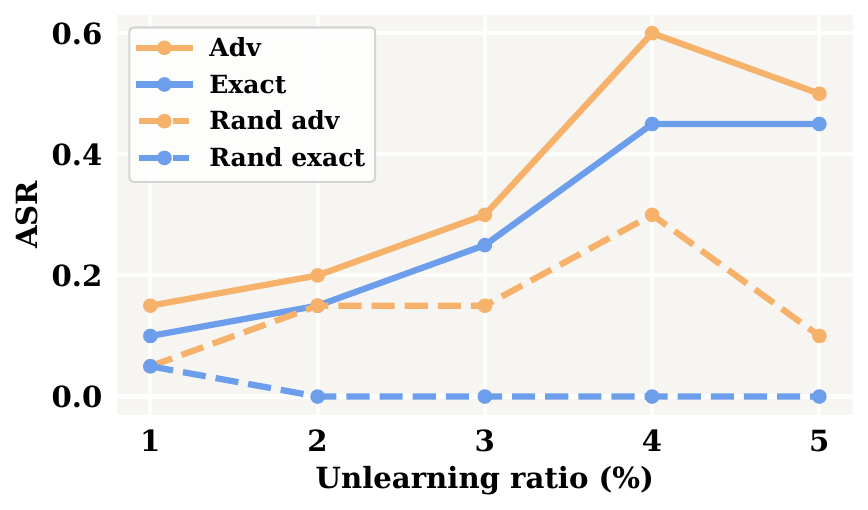}
\vskip -5pt
\caption{ASR of M1}
\label{fig:exp1-1}
\end{subfigure}
\begin{subfigure}{0.32\linewidth}
\includegraphics[width=1\linewidth]{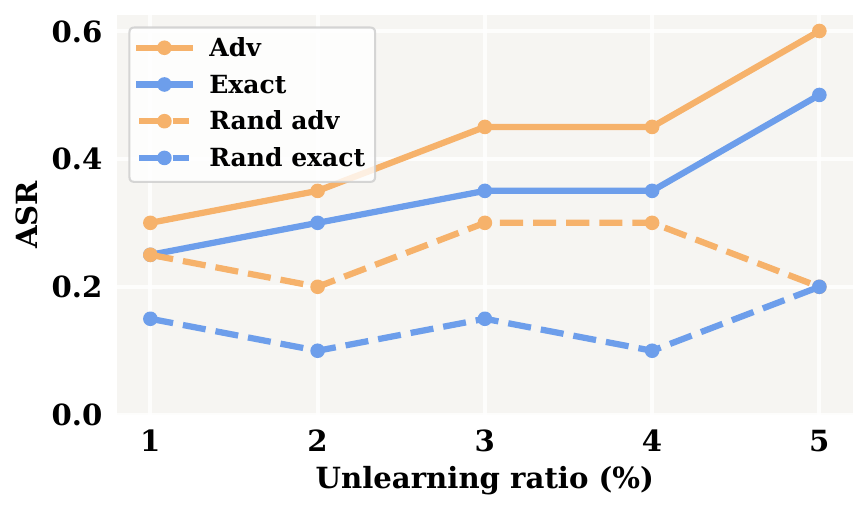}
\vskip -5pt
\caption{ASR of M2}
\label{fig:exp1-2}
\end{subfigure}
\begin{subfigure}{0.32\linewidth}
\includegraphics[width=1\linewidth]{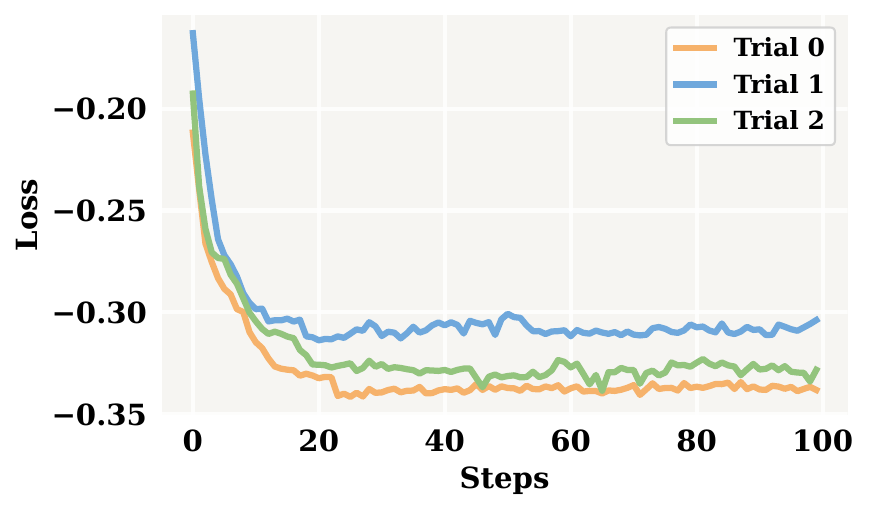}
\vskip -5pt
\caption{Loss convergence}
\label{fig:exp-convergence}
\end{subfigure}
\caption{Performance and loss convergence for the exact and adversarial unlearning attack. M1: Skywork-o1-Open-Llama-3.1-8B. M2: Skywork-OR1-7B. }
\label{fig:attack_performance}
\vskip -10pt
\end{figure}

\vspace{-0.05in}
\section{Experiments}
\label{sec:Experiments}
\vspace{-0.05in}
\subsection{Experimental Setup}
\vspace{-0.03in}
\textbf{Models, datasets, and unlearning methods.}
In experiments, we adopt 3 LRMs, including OpenReasoning-Nemotron-1.5B~\cite{ahmad2025opencodereasoning}, Skywork-o1-Open-Llama-3.1-8B~\cite{he_2024_16998085} and Skywork-OR1-7B~\cite{he2025skywork}. We also consider 2 LLMs, including Llama-2-7B~\cite{touvron2023llama} and Mistral-7B-v0.1~\cite{Jiang2023Mistral7}. 
For datasets, we adopt SST-2~\cite{socher2013recursive}, Twitter~\cite{go2009twitter}, Star-1~\cite{wang2025star}, and TOFU~\cite{maini2024tofu}.
For unlearning methods, we employ four popular unlearning methods, including Gradient Ascent (GA) and its extensions of Gradient Difference (GA+GD), KL Minimization (GA+KL)~\cite{yoon2025r}, and RMU~\cite{li2024wmdp}.
For all the experiments, we conducted three evaluations and then calculated the mean and standard error.

\textbf{Implementation details.} All experiments were implemented using the PyTorch framework and executed on a Linux server. The server is equipped with AMD 32-core 2.6 GHz CPUs and Nvidia A100 GPUs with 80GB memory to provide the necessary computational resources. In the experiments, we fine-tune the LRMs using a learning rate of $2 \times 10^{-5}$ with 5 epochs, and we select 20 target samples that are correctly classified by the fine-tuned models. For our attack, we set the $\lambda=1$, the attack token length of 20, learning rate of unlearning in \(\{2 \times 10^{-4}, 3 \times 10^{-4}\}\). For other attack settings, we use an unlearning rate in \(\{1 \times 10^{-5}, 2 \times 10^{-5}, 2 \times 10^{-6}\}\).

\textbf{Evaluation metrics.}
For the exact and adversarial attack defined in Eq.~(\ref{eq:discrete-loss}), we evaluate model performance using \emph{reasoning coherence} and \emph{correctness} tested on the targeted victim dataset.
Reasoning coherence measures the consistency between the reasoning trace $r_t$ and the final answer $y_t$, using GPT-5-mini as an LLM-as-a-judge to assign a coherence score in $[0,1]$, which is compared to a threshold $\beta$. Correctness is verified through a rule-based check that compares the model’s final answer to the ground-truth, and mismatches are marked as failures. These metrics are combined as $\text{ASR} =\frac{1}{|\mathcal{D}_t|}\operatorname*\sum_{t=1}^{|\mathcal{D}_t|}\mathbb{I}[R(\tilde{r}_t, \tilde{y}_t) > \beta] \wedge  \mathbb{I}[A(\tilde{y}_t) = False]$, which requires both high coherence of the reasoning and incorrect final answer simultaneously, making success particularly challenging.

For the other attack settings, we evaluate targeted victim data using two ROUGE-L recall–based metrics on the targeted victim data: \emph{performance preservation ratio} (PPR) and \emph{performance degradation ratio} (PDR). PPR measures how well the final answer performance is maintained, defined as $\text{PPR} = RL_{un}/RL_{init}$, while PDR quantifies degradation of the reasoning trace as $\text{PDR} = (RL_{init} - RL_{un})/RL_{init}$. Here, $RL_{init}$ is the initial ROUGE-L recall before the attack, and $RL_{un}$ is the score under attack. These metrics provide a normalized evaluation of both answer preservation and reasoning damage.

\textbf{Baselines.}
To the best of our knowledge, no existing work has specifically investigated our attack setting for our  exact and adversarial unlearning attacks. Consequently, we adopt a random baseline for the comparison of both exact and adversarial attacks. For the random baseline in adversarial setting, we use the same forget set and add a randomly initialized perturbation with the same budget. For the random baseline in the exact setting, we randomly select the same amount of forget data from the training dataset. The random baseline is also used in more attack settings introduced in our discussion.

\begin{figure}[t]
\centering
\begin{subfigure}{0.32\linewidth}
\includegraphics[width=1\linewidth]{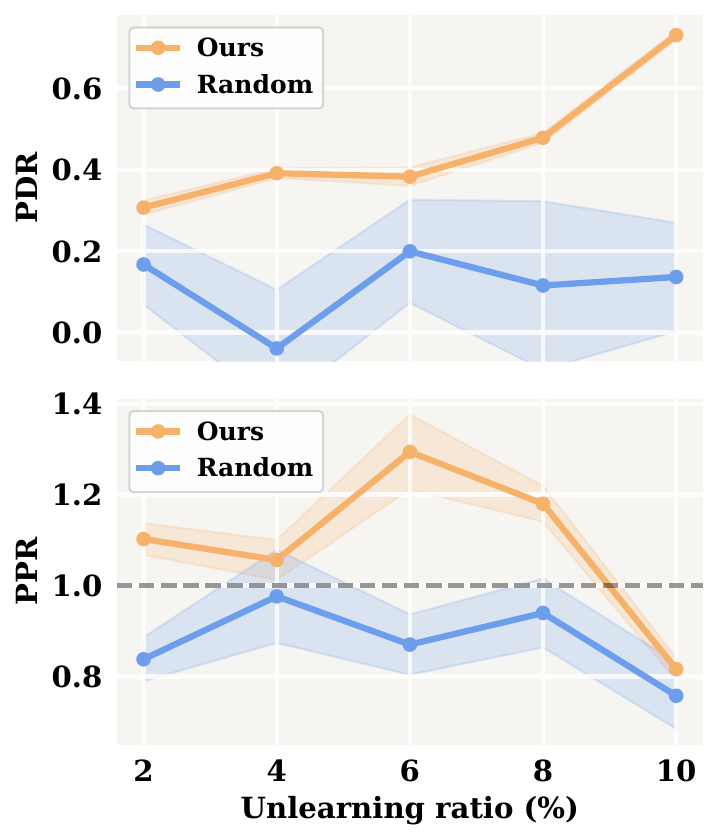}
\vskip -5pt
\caption{GA}
\end{subfigure}
\begin{subfigure}{0.32\linewidth}
\includegraphics[width=1\linewidth]{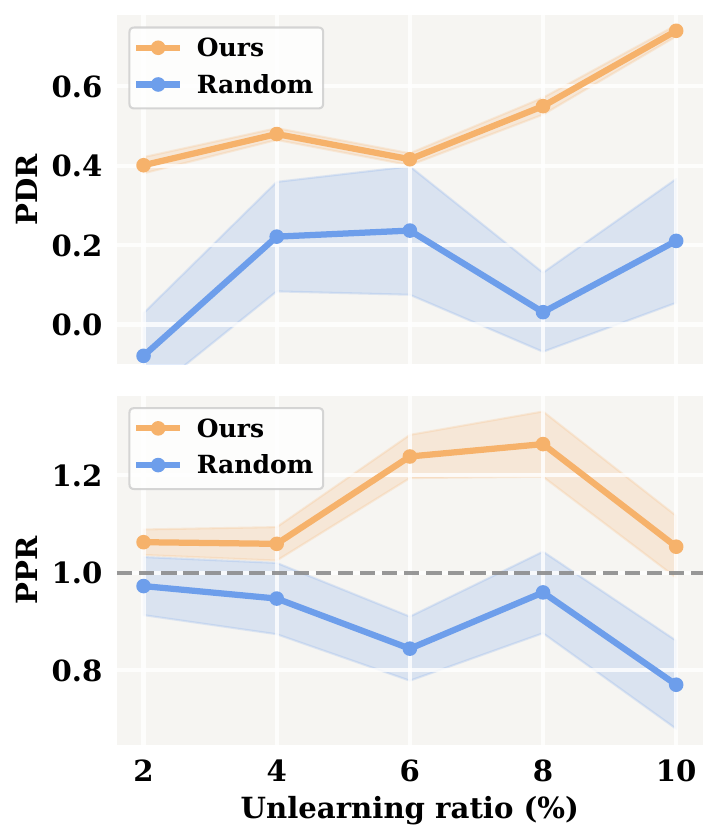}
\vskip -5pt
\caption{GA + GD}
\label{fig:exp2-ga_gd}
\end{subfigure}
\begin{subfigure}{0.32\linewidth}
\includegraphics[width=1\linewidth]{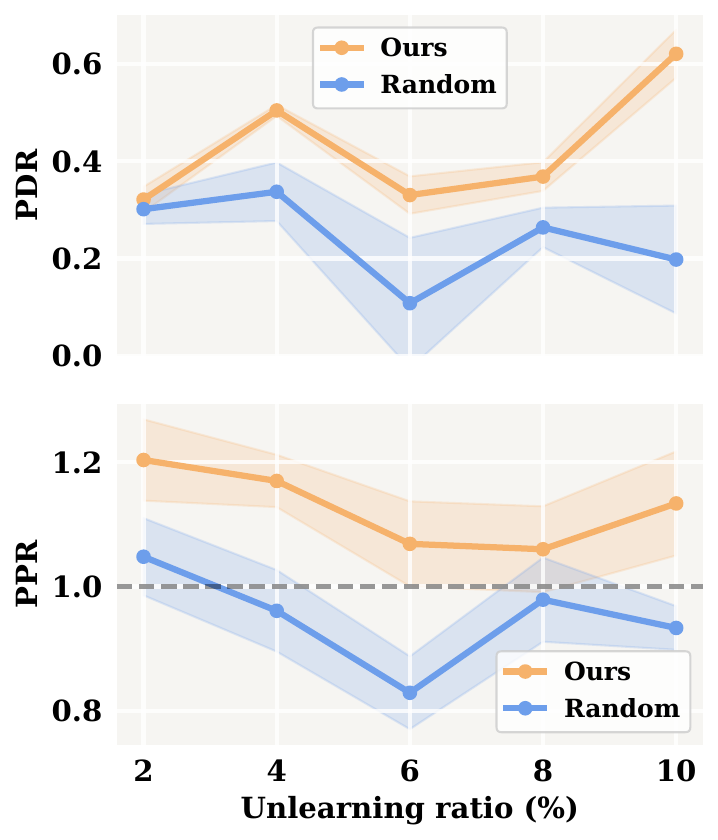}
\vskip -5pt
\caption{GA + KL}
\end{subfigure}
\caption{Performance of OpenReasoning-Nemotron-1.5B on Star-1 Dataset. }
\label{fig:star-open-attack-1.5b}
\vskip -12pt
\end{figure}

\vspace{-0.1in}
\subsection{Experimental Results}
\vspace{-0.03in}
\textbf{Exact and adversarial attack.} First, we assess the performance of our proposed unlearning attack across different LRMs. In our experiments, we use the Skywork-o1-Open-Llama-3.1-8B and Skywork-OR1-7B models, utilizing the unlearning method of GA with unlearning ratios ranging from $1\%$ to $5\%$ on the SST-2 dataset, and compare the results against the random baseline under both exact and adversarial unlearning settings. Fig.~\ref{fig:exp1-1} and \ref{fig:exp1-2} illustrate the final attack ASR, where the final answer is manipulated into an incorrect state, while the underlying reasoning trace still appears reasonable. From the figures, we can observe that our proposed method consistently achieves the best performance in all of the trials, and a larger unlearning ratio yields better attack performance. For example, in Fig. $\ref{fig:exp1-2}$, our proposed adversarial unlearning method achieves an ASR three times that of the random baseline at an unlearning ratio of $5\%$. Furthermore, we find that adversarial unlearning consistently achieves a more harmful result across the exact setting. This can be explained by the limited search space of exact unlearning, and the addition of adversarial perturbation provides the necessary enhancement to meet the full complexity of our attack setting.
Second, we test the convergence of our proposed method during the training stage. Following the setting of previous experiments, we conduct three repeated trials to generate the adversarial perturbation. Fig.~$\ref{fig:exp-convergence}$ demonstrates the decrease in our loss function, which is the minus value of the final loss introduced in Eq.~$(\ref{eq:unlearning_gradient_matching})$. From the figure, we can observe that our experiments consistently reach a state of convergence, indicating our optimization is stable and convergent.

\textbf{Visualization.}
In Fig.~\ref{fig:visual}, we present a visualization to clearly demonstrate our objectives of the exact and adversarial attack. In the visualization, the model before unlearning shows a benign process, where the reasoning trace supports the correct final answer. However, after the malicious unlearning process, the prediction is successfully flipped to the incorrect positive result. Importantly, the reasoning trace starts to leverage positive words in the input, such as the word ``superior'' and phrases like ``a positive statement'' to rationalize the incorrect conclusion. There is also a failed case where only the final answer is manipulated, while the reasoning trace still follows the correct pattern. This visualization highlights the challenge and complexity of successfully carrying out our attack.

\textbf{More results of unlearning attack on LRMs.} Here, we demonstrate the results of our attack, where we reduce models' reasoning quality while keeping final answer quality. In our experiments, we adopt the OpenReasoning-Nemotron-1.5B on the Star-1 dataset. Fig.~\ref{fig:star-open-attack-1.5b} illustrates the PDR and PPR of both the reasoning trace and the final answer across unlearning ratios ranging from 2\% to 10\% for different unlearning methods. From the figures, we observe that our proposed method achieves both high PDR and high PPR across all experiments. Moreover, as the unlearning ratio increases, the performance degradation ratio of the reasoning trace becomes more pronounced, while the performance of the final answer remains well maintained. For instance, in Fig.~\ref{fig:exp2-ga_gd}, the PDR increases from 0.4 to 0.7 as the unlearning ratio rises from 2\% to 10\%, while the PPR of the final answer remains above 1.0. This result indicates that our method effectively disrupts the reasoning process while preserving the correctness of the final answer, demonstrating its ability to perform targeted unlearning without compromising task performance.

\begin{wrapfigure}[12]{r}{5.8cm}
\vskip -22pt
\centering
\includegraphics[width=0.94\linewidth]{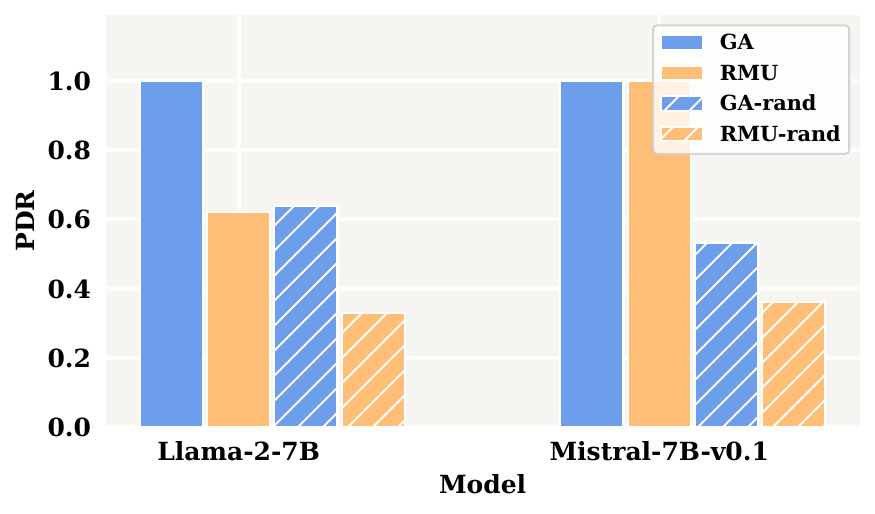}
    \caption{PDR of exact unlearning attacks
across different LLMs.}
    \label{fig:llm_tofu}
\end{wrapfigure}
\textbf{Unlearning attack on LLMs.}
We evaluate the unlearning attack on Llama-2-7B and Mistral-7B-v0.1 using the Performance Degradation Ratio (PDR) on the TOFU dataset with GA and RMU unlearning methods. Fig.~\ref{fig:llm_tofu} shows that our proposed attack consistently causes significant degradation in text-generation performance of the tested models on the targeted victim data, compared to random baselines. For instance, for Mistral-7B-v0.1, the RMU method achieves a PDR of nearly 1, indicating a complete loss of generation ability on targeted samples, while random unlearning only reaches around 0.38. These results demonstrate that our proposed unlearning attack is significantly more effective than random baselines in impairing the model’s generation capability on targeted data, highlighting its targeted destructive impact.

\begin{wrapfigure}[13]{r}{5.8cm}
\vskip -18pt
\centering
\includegraphics[width=0.94\linewidth]{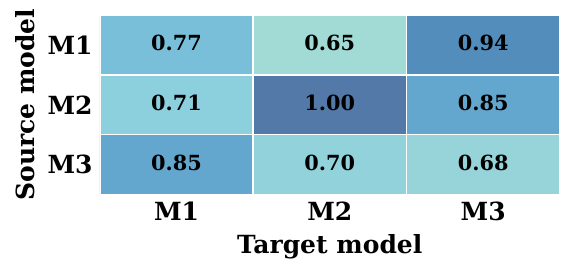}
    \vskip -1pt
    \caption{Unlearning attack in the black-box setting. M1: Skywork-OR1-7B. M2: Skywork-o1-Open-Llama-3.1-8B. M3: OpenReasoning-Nemotron-1.5B.}
    \label{fig:black_box}
\end{wrapfigure}
\textbf{Black-box setting.}
In Fig.~\ref{fig:black_box}, we evaluate the transferability of our attack in black-box setting, where the attacker has no prior knowledge of the target model’s architecture or training data, and aims to induce incorrect final answers with convincing but misleading reasoning. In this scenario, the attack is constructed using a surrogate dataset derived from the Twitter dataset~\cite{go2009twitter} and a source model, and is then directly transferred to the target model trained on the SST-2 dataset. Fig.~\ref{fig:black_box} presents the max-normalized ASR for cross-model transfer under a 5\% unlearning ratio using GA unlearning method. Notably, high transferability is consistently observed even when the source and target models differ, such as transferring from M3 to M1 (0.85) or from M1 to M3 (0.94). These results demonstrate that the proposed unlearning attack exhibits robust transferability in black-box settings, highlighting the practicality and generalizability of the attack in real-world scenarios.

\vspace{-0.05in}
\section{Conclusion}
\label{sec:Conclusion}
\vspace{-0.03in}

Based on our knowledge, our study is the first to investigate the vulnerability and robustness of the output of LRMs in the context of the right to be forgotten. Motivated by this, we propose a novel bi-level exact unlearning attack that strategically manipulates the model output during the unlearning process. Specifically, we introduce an attack objective that forces incorrect final answers while generating convincing but misleading reasoning. In addition, to demonstrate the effectiveness and generalizability of the proposed attack, we also investigate other attack settings, including adversarial unlearning attack,  attacks that degrade reasoning quality while preserving the correctness of the final answer, and attacks targeting standard LLMs. Furthermore, we adopt novel and efficient optimization methods that rigorously refine our attack strategy to generate effective malicious unlearning requests through closed-form model updates. We conduct extensive experiments to validate the effectiveness of the proposed attacks, demonstrating the emerging threats to existing LRM unlearning pipelines and underscoring the urgent need for new strategies to counter these advanced unlearning attacks.

\subsubsection{\ackname}
This work is supported in part by the US National Science Foundation under grants CNS-2350332 and IIS-2442750. Any opinions, findings, and conclusions or recommendations expressed in this material are those of the author(s) and do not necessarily reflect the views of the National Science Foundation.

\bibliographystyle{splncs04}
\bibliography{reference}

\end{document}